\documentclass[letterpaper]{article} 
\usepackage{aaai23}
\usepackage{natbib}  
\usepackage{times}  
\usepackage{helvet}  
\usepackage{courier}  
\usepackage[hyphens]{url}  
\usepackage{graphicx} 
\usepackage{amsmath}
\usepackage{graphicx} 
\usepackage{caption} 
\usepackage{algorithm}
\usepackage{algpseudocode}

\usepackage{newfloat}
\usepackage{listings}
\DeclareCaptionStyle{ruled}{labelfont=normalfont,labelsep=colon,strut=off} 
\floatstyle{ruled}
\newfloat{listing}{tb}{lst}{}
\floatname{listing}{Listing}

\newcommand{\etal}{\textit{et al.}}
\newcommand{\eg}{\textit{e.g.,}}

\newcommand{\metricname}{Deferred Error Volume}
\newcommand{\metricnameshort}{DEV}
\newcommand{\depthconstraintshort}{DDC}
\newcommand{\rateconstraintshort}{DR}
\title{Evaluating and Improving Interactions with Hazy Oracles}
\author {
    Stephan J. Lemmer and 
    Jason J. Corso 
}
\affiliations {
    University of Michigan, Ann Arbor\\
    500 S State St, Ann Arbor, MI 48109\\
    lemmersj@umich.edu, jjcorso@umich.edu
}
\begin{document}

\maketitle

\begin{abstract}
Many AI systems integrate sensor inputs, world knowledge, and human-provided information to perform inference. While such systems often treat the human input as flawless, humans are better thought of as \textit{hazy oracles} whose input may be ambiguous or outside of the AI system's understanding. In such situations it makes sense for the AI system to defer its inference while it disambiguates the human-provided information by, for example, asking the human to rephrase the query. Though this approach has been considered in the past, current work is typically limited to application-specific methods and non-standardized human experiments. We instead introduce and formalize a general notion of deferred inference. Using this formulation, we then propose a novel evaluation centered around the \metricname{} (\metricnameshort{}) metric, which explicitly considers the tradeoff between error reduction and the additional human effort required to achieve it. We demonstrate this new formalization and an innovative deferred inference method on the disparate tasks of Single-Target Video Object Tracking and Referring Expression Comprehension, ultimately reducing error by up to 48\% without any change to the underlying model or its parameters.
\end{abstract}

\section{Introduction}
Many artificial intelligence systems are motivated by intuitive interaction with humans: by combining sensor inputs, world knowledge, and human-provided information, they perform useful inferences such as answering visual questions~\cite{antol_vqa_2015}, propagating an initial segmentation through a video~\cite{perazzi_benchmark_2016}, or integrating semantic information to resolve perceptual ambiguity~\cite{szeto_click_2017}. 
\begin{figure}[t]
\centering
\includegraphics[width=\linewidth]{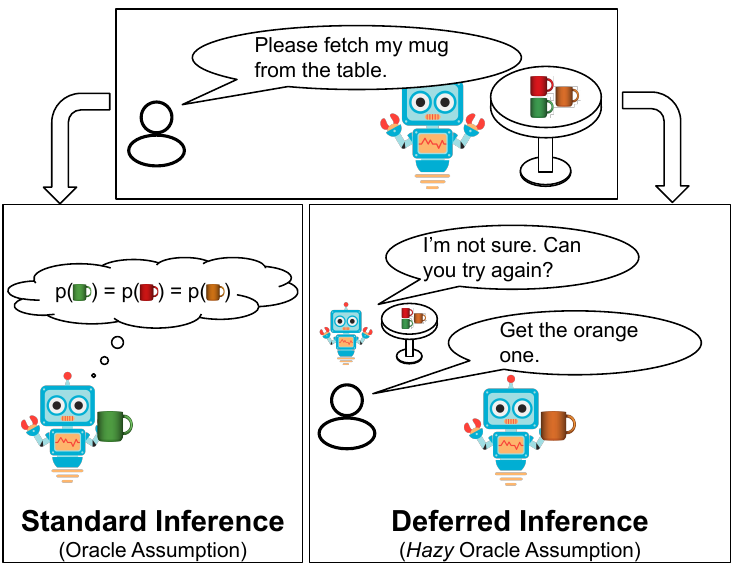}
\caption{A simple example of the benefit of our approach. The human provides an ambiguous request that standard inference will misinterpret two-thirds of the time. By assuming a hazy oracle, the robot is able to defer the inference and obtain additional information to solve the task. Robot image via Wikimedia commons (CC-SA), mug images via freepik.com. Best viewed in color.}
\label{fig:ledefigure}
\end{figure}
Despite the well-understood ambiguities and difficulties of human-provided information explored in the human-computer interaction domain~\cite{ipeirotis_quality_2010,bhattacharya_why_2019,gordon_disagreement_2021}, state-of-the-art works in machine learning generally treat the human as an oracle that provides a single piece of flawless information then disappears. While this formulation simplifies dataset-based evaluation, we believe it more appropriate to treat humans as \textit{hazy oracles}: the information they provide may be incorrect, ambiguous, or outside of the system's understanding of the world, but they can provide clarifying information after the initial query.

For example, consider the scenario shown in Figure~\ref{fig:ledefigure}: a robot must perform the task of retrieving an orange mug given a verbal command by a user. While the robot is likely to misunderstand \textit{please fetch my mug from the table} if it does not know which mug belongs to the speaker, it can accomplish the desired task by delaying its response and requesting more information from the human. While conceptually simple, this mechanism of \textit{deferred inference} is a challenge in practice: the robot must determine whether to defer and, if it chooses to defer, determine how best to integrate the potentially noisy additional information.

Although the benefit of deferred inference is intuitive both for individual~\cite{gurari_vizwiz_2018} and crowd~\cite{lemmer_crowdsourcing_2021} interactions, no standard approach for implementation or evaluation has emerged. Selective prediction approaches~\cite{chow_optimum_1970,geifman_selective_2017} evaluate performance thoroughly but do not consider subsequent inputs for the same task, and methods that defer inference~\cite{mees_composing_2020,sharma_correcting_2022} typically validate their methods via small-scale problem-specific human experiments, limiting dataset size and comparability between evaluations. Additionally, since such studies require deferral criteria to be set prior to the experiment, the complex interplay between accuracy and burden on the user is only described at one point. This is a critical oversight, as the best deferral method often changes based on how many deferrals are permitted~\cite{lemmer_crowdsourcing_2021,lemmer_ground-truth_2021}.

To appropriately facilitate interaction with hazy oracles, we therefore introduce a more principled formulation that addresses these limitations and enables low-cost direct-comparison between deferred inference methods. To do this, we propose a new metric for measuring performance---\metricname{} (\metricnameshort{})---that explicitly and thoroughly evaluates the tradeoff between inference error and the additional human effort required to achieve it. Alongside this evaluation method, we introduce a method for deferred inference based on a belief update. While simple, our proposed method is both generalizable and remarkably effective: it outperforms previous methods by more effectively integrating additional information and reducing the likelihood of multiple deferrals, and it can be easily applied to novel applications and architectures.

We demonstrate this versatility by implementing our evaluation and method on the disparate applications of Single-Target Video Object Tracking~\cite{kristan_novel_2016} and Referring Expression Comprehension~\cite{mao_generation_2016}. On both applications, we show significant improvement over both deferral-free inference and deferral methods proposed in other works, reducing error by up to 48\% under an acceptable level of human effort with no modifications to the training or architecture of the underlying task model.

Our main contributions are as follows:
\begin{itemize}
    \item A general formulation and evaluation method for deferred inference with hazy oracles, centered around the \metricname{} (\metricnameshort{}) metric, that provides a fair and thorough comparison between methods.
    \item A deferred inference method that significantly improves performance over both the base model without deferral and deferral baselines based on previous work.
    \item An Evaluation of our method on the disparate tasks of Visual Object Tracking and Referring Expression Comprehension that shows the need for a novel evaluation, the generalizability of our evaluation and solution, and the quantitative benefit---up to a 48\% reduction in error---of our proposed method.
\end{itemize}


\section{Related Work}
\label{sec:related_work}
\paragraph{Aggregating Human Inputs}
Many works, particularly in the crowdsourcing domain, use multiple human inputs to increase accuracy. Though some works \cite{branson_visual_2010,russakovsky_best_2015} allow the model to choose when to terminate, the most common approach is to allow the human operator to review the model's output directly and provide new information until the result is satisfactory~\cite{jain_click_2016,gouravajhala_eureca_2018,choi_aila_2019,agustsson_interactive_2019,uijlings_panoptic_2020}. These approaches are sufficient for dataset collections: performing tasks such as answering questions about given bounding boxes~\cite{russakovsky_best_2015} or confirming answers~\cite{uijlings_learning_2018} is faster and more accurate than generating the dataset through drawing a bounding box directly on the image. However, such methods encode the assumption that one can interact directly in the output space (\eg{} you are capable of understanding and manipulating the image). This assumption is impractical in important cases: if a visual question answering system is assisting a visually impaired individual~\cite{gurari_vizwiz_2018}, that individual can not manually confirm if the answer is correct, but can easily rephrase the question.

\paragraph{Deferred Inference} Deferred inference can be meaningfully applied to many applications~\cite{antol_vqa_2015, das_visual_2017,szeto_click_2017,anderson_vision-and-language_2018,perazzi_benchmark_2016}, but the majority of existing works focus on Visual Question Answering~\cite{mahendru_promise_2017,uehara_learning_2022} or Referring Expression Comprehension~\cite{nyga_grounding_2018,shridhar_interactive_2018,mees_composing_2020,sharma_correcting_2022}, likely due to the intuition of asking for more information during conversation. Such methods---which often rely on a difficult text-generation task---require some combination of restrictive assumptions, more complex architectures, novel datasets, and human experimentation. Two works~\cite{hatori_interactively_2018,lemmer_crowdsourcing_2021} follow our approach and mitigate these challenges by receiving a deferral response in the same format as the task definition. These works serve as baselines for both our evaluation and proposed deferral method.

\paragraph{Evaluation Methods}
Most works related to deferred inference set deferral conditions prior to their experiments and report the final error. When user burden is specified, it is most often a point metric corresponding to the reported accuracy, such as time-per-annotation~\cite{agustsson_interactive_2019,uijlings_panoptic_2020} or number of annotations per target~\cite{ipeirotis_quality_2010,hatori_interactively_2018}. The value of these point measurements is questionable: when a deferral method is evaluated at different thresholds, the best method often changes~\cite{lemmer_ground-truth_2021,lemmer_crowdsourcing_2021}.

Works in selective prediction~\cite{chow_optimum_1970,cortes_boosting_2016,geifman_bias-reduced_2019,yildirim_leveraging_2019,mozannar_consistent_2020} often evaluate a relevant tradeoff between the proportion of classifications performed and the error, but implicitly or explicitly assume the human can provide the correct answer---which can not be guaranteed in settings such as crowdsourcing~\cite{lemmer_crowdsourcing_2021} or public-facing deployments~\cite{bhattacharya_why_2019}. An exception to this is the work of Bondi \etal{}~\cite{bondi_role_2022}, which acknowledges the possibility of human failure after deferral, but, like crowdsourcing works, requires the human to directly perform the task.

\section{Problem Statement}
An automated agent is asked to perform a task, such as cropping an image based on a language request or tracking an object through a video. This agent has some probability of solving the task on its own, but may also defer to a \textit{hazy oracle} that can provide additional information at some cost. The information provided by the hazy oracle may be ambiguous either in truth (\eg{} more than one output satisfies the request) or because it is outside of the agent's understanding of the world (\eg{} a model trained in English will not understand a Spanish query, regardless of the information it contains). With the goal of minimizing error subject to constraints on human effort or human effort subject to constraints on error, the agent must determine whether to defer its decision and request information from the hazy oracle. If the agent chooses to defer its decision, it must additionally determine how best to integrate the additional, potentially noisy, information provided. 

\begin{figure}[t]
\centering
\includegraphics[width=\linewidth]{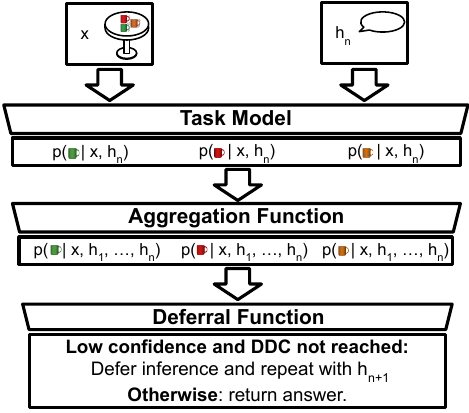}
\caption{Our formulation of deferred inference, with the task shown in Figure~\ref{fig:ledefigure} used for illustration. Here, $x$ is the robot's perception and $h_n$ is the human input at step $n$.}
\label{fig:rq_method}
\end{figure}
\begin{table}[b]
\centering
\includegraphics[width=\linewidth]{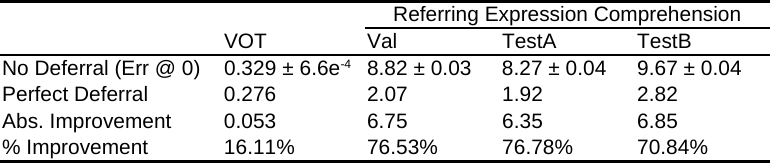}
\caption{Error for the task model with No Deferral (Err @ 0) and Perfect Deferral on two applications. A perfect deferral method can reduce error by over 76\%. Err @ 0 is reported as mean and standard error of 100 trials. Full setup described under Exemplar Applications.}
\label{tab:motivation_table}
\end{table}
We show an abstraction of such an agent, modifying the terminology of previous work~\cite{lemmer_crowdsourcing_2021}, in Figure~\ref{fig:rq_method}. The abstraction consists of three parts: the \textit{task model} produces a prediction based on the input data, the \textit{aggregation function} accepts one or more predictions corresponding to different human inputs from the task model and produces a combined prediction, and the \textit{deferral function}---which consists of a \textit{deferral score} and a threshold---determines whether or not a deferral should occur. While we discuss these as distinct entities, this is not a requirement: a recurrent neural network with a deferral score similar to SelectiveNet~\cite{geifman_selectivenet_2019} would execute all three functions in a single step.
\begin{figure}[t]
\centering
\includegraphics[width=\linewidth]{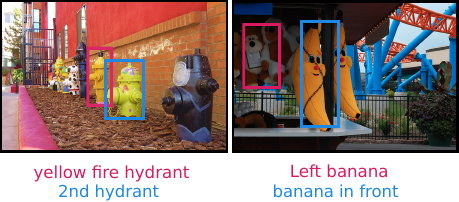}
\caption{Whether a phrase is semantically ambiguous (left) or simply unclear to the model (right), a new human input can change an inference from incorrect (pink) to correct (blue). Best viewed in color.}
\label{fig:textmatters}
\end{figure}
%
%
%
\paragraph{Motivation}
We motivate the problem through the applications of single-target video object tracking, where the goal is to propagate bounding box drawn around an object in the first frame through all subsequent frames, and referring expression comprehension, where the goal is to draw a bounding box around the object described by a text query. 
We show the benefit of a perfect deferred inference method---that is, one that can select the best human input from the dataset---quantitatively in Table~\ref{tab:motivation_table}: for the validation split of the referring expression comprehension task, \textit{using the best human input can reduce error by over 76\%}.

In Figure~\ref{fig:textmatters} we demonstrate the benefit qualitatively by showing four human inputs and their matching outputs on the application of referring expression comprehension. On the left we see the more intuitive case, where the first expression, \textit{yellow fire hydrant}, can be reasonably thought to refer to four objects, while \textit{2nd hydrant} is mostly unambiguous.\footnote{While data collection was designed to produce unambiguous referring expressions, we find a non-negligible number of semantically ambiguous examples. This is likely due to the requirement that annotators make a guess for every expression.} On the right, we see a case where the referring expression \textit{left banana} isn't truly ambiguous, but the model produces the wrong answer due to shortcomings in its understanding of language. Though the latter failure is more commonly considered a shortcoming of the model, the fact that a new referring expression can successfully solve both tasks demonstrates the potential benefit of deferred inference with modern, imperfect, models.

\section{Proposed Evaluation}
\paragraph{\metricname} The exact performance of a deferred inference method is best measured as a combination of three different factors: i) the error, which is a property of the application; ii) the Deferral Rate (\rateconstraintshort{}), which is the expected number of deferrals that will occur for each task; and, iii) the Deferral Depth Constraint (\depthconstraintshort{}), which is the maximum number of times that a task can be deferred. Since evaluating at only a single \rateconstraintshort{}-\depthconstraintshort{} pair does not provide an adequate analysis of a deferral method, we propose the \metricname{} (\metricnameshort{}), which finds the error at every potential combination of \rateconstraintshort{} and \depthconstraintshort{} then calculates the volume under that surface. 

While both \rateconstraintshort{} and \depthconstraintshort{} are theoretically unconstrained, calculating the volume under a surface requires bounds to be placed. To produce these bounds, we make the least restrictive assumptions possible: the deferred inference method is capable of deferring every task at least once and the deferral function consists of a deferral score followed by a threshold. The former places an upper bound on \rateconstraintshort{} at one and a lower bound on \depthconstraintshort{} at one, and the latter allows a thorough evaluation of the relationship between \rateconstraintshort{} and error. We set an upper bound of ten on the \depthconstraintshort{}, which captures all practical deferral depths, and divide by ten to scale the width of this dimension to one. We discuss the implications of this upper bound in our results.

\begin{algorithm}[t]
\caption{Calculating \metricnameshort{}}
\label{alg:rev}
\begin{algorithmic}
\State \metricnameshort{} $\gets$ 0
\State \depthconstraintshort{} $\gets$ 1
\While{\depthconstraintshort{} $\leq$ 10}
\State tasks $\gets$ draw\_tasks()
\State \metricnameshort{} $\gets \text{\metricnameshort{}}+\frac{\text{calc\_error(tasks})}{10(\text{len(tasks)}+1)}$
\State N $\gets$ 0
\While{N $ < $ len(tasks)}
    \State cur\_task $\gets$ find\_task\_to\_defer(tasks, \depthconstraintshort{})
    \State response $\gets$ get\_new\_input(cur\_task)
    \State updated\_task $\gets$ aggregate\_fn(cur\_task, response)
    \State update\_tasks(tasks, updated\_task)
    \State \metricnameshort{} $\gets \text{\metricnameshort{}}+\frac{\text{calc\_error(tasks)}}{10(\text{len(tasks)}+1)}$
    \State N $\gets$ N + 1
\EndWhile
\State $\text{\depthconstraintshort{}} \gets \text{\depthconstraintshort{}} + 1$
\EndWhile
\end{algorithmic}
\end{algorithm}

Since evaluation will be performed on finite datasets, the volume under the curve when using rectangular integration is the mean of error under all constraint sets:
\begin{align}
    \text{DEV} = \frac{1}{10(N+1)}\sum\limits_{\text{\depthconstraintshort{}}=1}^{10}\sum\limits_{n=0}^{N} \ell(\rateconstraintshort=\frac{n}{N}, \depthconstraintshort{})\enspace,
\end{align}
\noindent where $\ell(\rateconstraintshort{},\depthconstraintshort{})$ is the error at a specific \rateconstraintshort{} and \depthconstraintshort{}, and $N$ is the number of tasks in the dataset. We show the calculation of DEV in Algorithm~\ref{alg:rev}: after an initial error calculation with one randomly drawn human input for every task ($\textsf{\small draw\_tasks}$), $\textsf{\small find\_task\_to\_defer}$ finds the highest deferral score where the \depthconstraintshort{} constraint is not exceeded, draws another human input from the dataset ($\textsf{\small get\_new\_input}$), uses the aggregation function to update the prediction, updates the \metricnameshort{}, and repeats the process.

Such a thorough dataset-based evaluation has only one major requirement: \textit{there must be a method by which deferral responses can be provided}. This requirement can be satisfied in a number of ways: the deferral response may be of the same form as the initial piece of human information with a dataset containing multiple pieces of human input per task (the approach of this work), the deferral query and response may be from a set of pre-defined attributes~\cite{branson_visual_2010}, or an external agent capable of answering additional queries---potentially with access to oracle knowledge---may be developed~\cite{uehara_learning_2022}. 
\paragraph{Marginals}
To illustrate the effect of individual constraints on a method's performance, we marginalize out the \rateconstraintshort{} and \depthconstraintshort{} constraints and plot the result, referring to the measurement as \textit{mean error}. Notably, calculating these marginals requires no inferences beyond those already used to calculate the \metricnameshort{}.
\paragraph{Error}
To provide an additional intuitive measure of the performance improvement enabled by a deferral method, we report error at two specific locations: deferral rate of zero (Err @ 0), which is the base error of the task model, and deferral rate of one (Err @ 1), which corresponds to the error when the number of deferrals is equal to the number of tasks (if \depthconstraintshort{} $>$ 1, this does not mean every task will have exactly one deferral). Since these errors correspond to the first and last points on the relevant marginal plot, no additional calculation is required to obtain them.

\section{Proposed Method}
We propose a straightforward method for deferred inference that can easily be applied to many state-of-the-art task models. Underlying our method are two assumptions: first, we assume the task model output is a distribution. This is trivial for a softmax output but may require additional consideration for other output formats~\cite{harakeh_bayesod_2020}. Second, we assume the deferral function is based on this distribution. With these assumptions, we can then use a belief update as the aggregation function. If we treat human inputs $h_1, ..., h_n$ as independent and represent the non-deferrable portion of the input (\eg{} an image) as $x$, the probability of a specific output, $y$, is:
\begin{align}
    p(y|x, h_1, ..., h_n) \propto \prod_{n=1}^{N} p(y |x, h_n)\enspace.
\end{align}

Though this formulation permits the deferral and its response to take many forms,  we treat the deferral as a request for the human operator to rephrase the initial input. This allows our formulation to be rapidly applied to new research in the machine learning space, as it does not require novel datasets or architectural changes.

\begin{figure}[t]
\centering
\includegraphics[width=\linewidth]{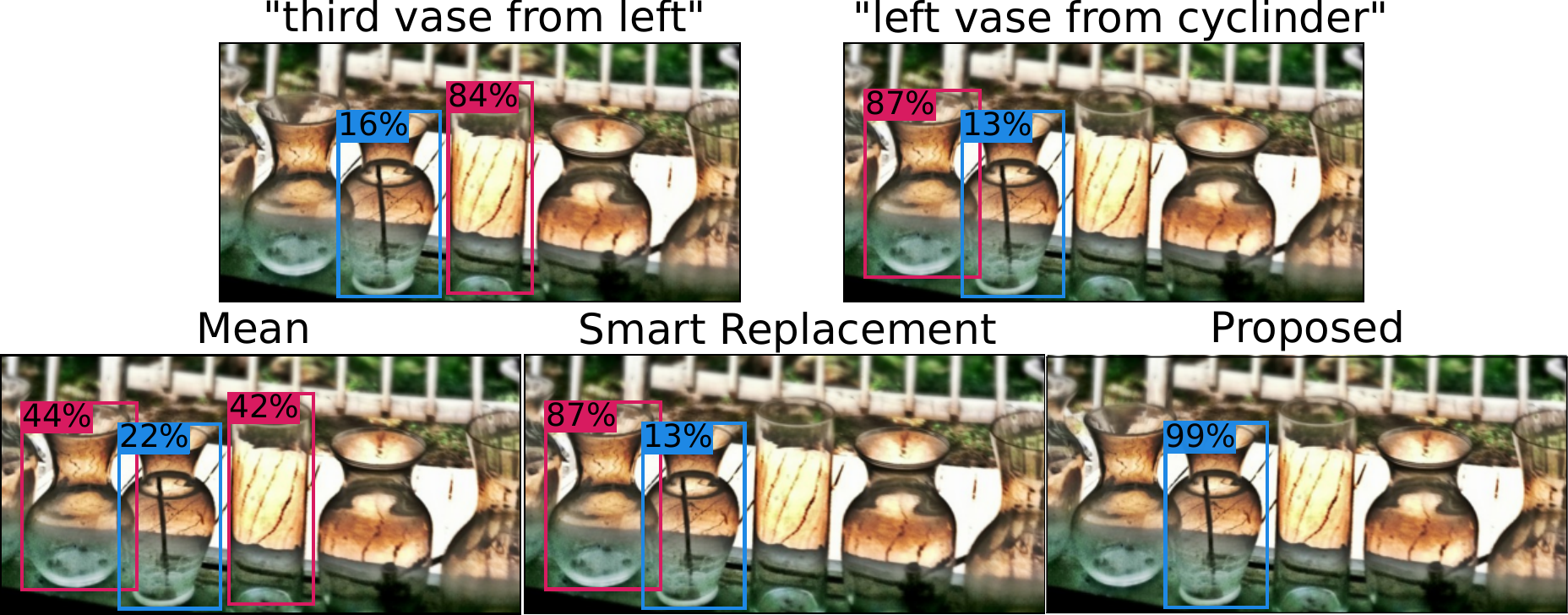}
\caption{Our proposed aggregation function quickly combines complementary information to achieve higher confidence and accuracy than previous methods such as taking the mean of two outputs~\cite{hatori_interactively_2018} or selecting the better output (Smart Replacement~\cite{lemmer_crowdsourcing_2021}). Target object boxed in blue. Original image cropped vertically for space. View in color.}
\label{fig:whycombine}
\end{figure}

Implementation of our formulation is illustrated intuitively for a classification task such as Referring Expression Comprehension (Figure~\ref{fig:whycombine}): an initial image, $x$, and referring expression, $h_1$, are given to the task model, resulting in a softmax output. If the deferral function decides that inference should be deferred based on this output, $h_2$ is solicited and passed through the task model alongside $x$, resulting in a second softmax output. The two output vectors are then multiplied elementwise, and the resulting vector is normalized. The deferral function is then executed on this normalized vector to determine if another deferral is necessary. As we see, combining the inferences in this way demonstrates a major benefit of our method: it quickly identifies the target object with high certainty, while aggregation functions such as using the better of the two outputs~\cite{lemmer_crowdsourcing_2021} or taking the mean of the two outputs~\cite{hatori_interactively_2018} would perform additional deferrals or return an incorrect answer depending on the given constraints.

\section{Exemplar Applications}
\subsection{Single-Target VOT}
\paragraph{Goal} In single-target Video Object Tracking (VOT), a human defines the task by drawing a bounding box around an object in the first frame of a video. Using only previous frames (the online tracking setting) the model then propagates this bounding box through the video. Despite the relatively low dimension of the input space, deferred inference is properly motivated for this application: driven by the sensitivity of tracking algorithms to their initializations, perturbation tests are a standard component of evaluation~\cite{wu_online_2013,kristan_novel_2016}.

\paragraph{Model and Dataset} Since it is the only VOT dataset, to our knowledge, that contains multiple annotations per tracked object, we perform our analysis using the crowdsourced data from Lemmer \etal{}~\cite{lemmer_crowdsourcing_2021}. This dataset consists of nine first-frame annotations for every video in the OTB-100 dataset~\cite{wu_online_2013}. To differentiate between low-quality inferences and inferences performed on an incorrectly selected object, we remove instances in this dataset where the initialization has an IoU of less than 0.5 with the gold-standard initialization. As our task model, we use the ToMP tracker~\cite{mayer_transforming_2022} with a ResNet-50 backbone~\cite{he_deep_2016} and weights provided by the original authors.

\paragraph{Base Error Metric} We measure performance using the mean intersection-over-union (IoU), which is commonly used in the evaluation of the VOT application~\cite{wu_online_2013, kristan_novel_2016}. To maintain the notion of error, we subtract the IoU from its maximum value of one. Unlike previous evaluations, we allow the model's inference to continue when the object track is lost: resetting the tracker requires a ground-truth box on every frame, which is intractable both in a real-world application and in the context of our implementation of deferred inference. 

\paragraph{Deferral Implementation} The output of ToMP is a single bounding box at every frame. To convert this to a distribution for our deferral and aggregation functions, we first produce stochastic bounding boxes by performing each inference 100 times with Monte Carlo dropout enabled~\cite{gal_dropout_2016} similar to previous work on object detection~\cite{harakeh_bayesod_2020}. Every bounding box is represented as a tuple of (TLX, TLY, Width, Height) and expectation maximization~\cite{dempster_maximum_1977} is performed on every frame to transform these representations into a Mixture of Gaussians. To determine the number of Gaussians for expectation maximization, we use DBSCAN~\cite{ester_density-based_1996} with epsilon 10 and minimum samples 20, which provided the lowest deferral-free error among the parameter set $\epsilon \in \{1, 3, 5, 10, 15, 20, 50\}$ and $\text{min\_samples} \in \{3, 5, 10, 15, 20\}$. We use Scikit-Learn~\cite{pedregosa_scikit-learn_2011} for both Expectation Maximization and DBSCAN.

Using these distributions, the deferral score is produced by randomly sampling 500 bounding box pairs and measuring the mean IoU between them. For both our method and baselines, we create an output bounding box by taking 10,000 samples from the mixture and using the one with the highest likelihood. For our method, these samples are scattered by adding a normally-distributed random value with standard deviation 7 to every dimension, which allows us to combine distributions that are close in Euclidean space but several standard deviations apart.
\begin{table*}[t]
\centering
\includegraphics[width=\linewidth]{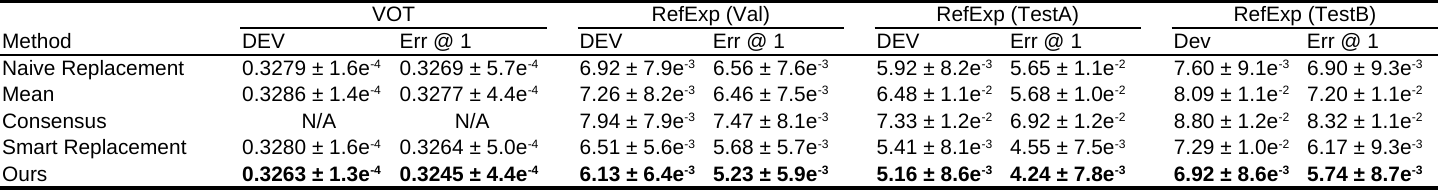}
\caption{The \metricnameshort{} and Err @ 1 metrics for baselines and our method (Err @ 0 shown in Table~\ref{tab:motivation_table}). Our method performs best across all applications and splits by a significant margin.}
\label{tab:summary_results}
\end{table*}
\subsection{Referring Expression Comprehension}
\paragraph{Goal} In referring expression comprehension, a task is defined by an image and text query, and the task model draws a bounding box around the object described by the text. The high dimensionality of the input space means that there is much room for both semantic ambiguity and gaps in the task model's knowledge that can be corrected or clarified after a deferral. 

\paragraph{Model and Dataset} For the task model, our evaluation uses the UNITER architecture~\cite{chen_uniter_2020}, which formulates referring expression comprehension as classification over a set of externally-provided bounding boxes. We provide ground-truth detections for these bounding boxes to minimize the influence of an external object detector. We train and evaluate on the RefCOCO~\cite{kazemzadeh_referitgame_2014} dataset because it contains multiple references to all but one target object, which is substantially better than both the RefCOCO+~\cite{kazemzadeh_referitgame_2014}, and RefCOCOg~\cite{mao_generation_2016} datasets. Our model is trained on a single GeForce GTX Titan XP GPU using the training settings given by the original authors with a few small modifications: we use full precision floating point operations, adjust the batch size from 128 to 64, and accumulate gradients over two steps. 

\paragraph{Base Error Metric} Consistent with previous work~\cite{mao_generation_2016}, performance on this application is measured using an indicator function. Error is zero if the predicted bounding box has an IoU of greater than 0.5 with the ground-truth bounding box, and 100 otherwise. Aggregate error can then be interpreted as the percentage of tasks that are completed incorrectly. We maintain the val, testA, and testB splits from previous works~\cite{yu_modeling_2016}, but note our evaluation measures per-task performance instead of per-phrase performance, making it incorrect to directly compare our results to other evaluations.

\paragraph{Deferral Implementation} Because the UNITER referring expression comprehension model outputs a softmax distribution, we use entropy as our deferral score and implement our proposed aggregation function by multiplying and normalizing the outputs. We improve the output's ability to detect ambiguity by performing Monte Carlo dropout with 100 passes, matching the number of passes in the original work~\cite{gal_dropout_2016}. We discuss the importance of using MC dropout in our technical appendix. 
\section{Experiments}
\begin{figure*}[t]
\centering
\includegraphics[width=0.85\linewidth]{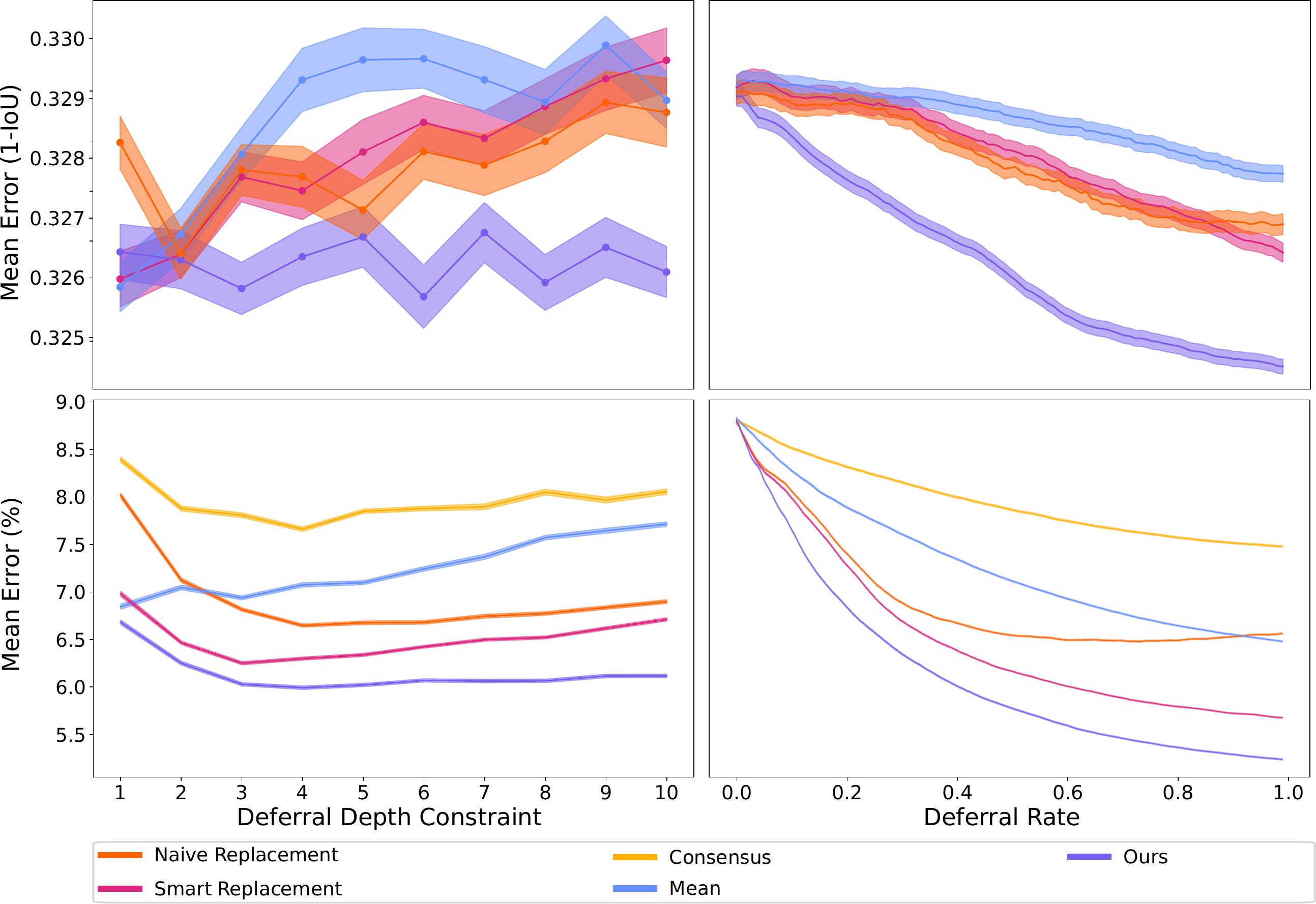}
\caption{Marginal plots showing the effect of the \depthconstraintshort{} (left) and \rateconstraintshort{} (right) on the VOT (top) and Referring Expression Comprehension (val split) (bottom) applications. Shaded area represents one standard error across 100 trials. View in color.}
\label{fig:marginals_both}
\end{figure*}
\paragraph{Baselines}
We compare our proposed method to four aggregation functions adapted from previous work:
\begin{itemize}
    \item Naive Replacement: If a deferral is performed, the most recent input is always used. If no \depthconstraintshort{} is specified this is equivalent to a selective prediction approach~\cite{chow_optimum_1970,geifman_selective_2017}, where the user must restart the task if the inference is declined.
    \item Mean: If the inference is deferred, the mean of \depthconstraintshort{} new inputs is used. This is equivalent to the aggregation function of Hatori \etal{}~\cite{hatori_interactively_2018}, who implicitly defined a \depthconstraintshort{} of one.
    \item Consensus: If a deferral is performed, \depthconstraintshort{} new inputs are taken and the consensus of all outputs is returned as the answer. If there is no consensus, an answer is chosen randomly from the potential outputs with equal occurrences. This is a basic approach often used in crowdsourcing~\cite{deng_imagenet_2009, song_popup_2019}. We do not implement this baseline on the VOT application due to the high number of potential outputs.
    \item Smart Replacement: If inference is deferred, the deferral score between all responses is compared, and the output corresponding to the best deferral score is used. As with the Mean baseline, we extend the original work~\cite{lemmer_crowdsourcing_2021} by allowing the \depthconstraintshort{} to be greater than one.
\end{itemize}

\paragraph{Results}
We see in Table~\ref{tab:summary_results} that deferred inference improves over the no deferral condition both on the mean (DEV) and at a deferral rate of one (err @ 1) for all methods and problem settings. In other words, any of the evaluated aggregation functions are better than the no deferral condition. Further, our proposed aggregation function outperforms all baselines in all settings on the evaluated metrics, and reduces error between the deferral-free condition (Table~\ref{tab:motivation_table}-No Deferral) and deferral rate 1 (Table~\ref{tab:summary_results}-Err @ 1): error decreases by 1.37\% for VOT, 40.7\% for RefExp-Val, 48.7\% for RefExp-TestA, and 40.6\% for RefExp-TestB. In other words, \textit{our method is effective on two very different applications, and can reduce error by over 48\% (from 8.27\% to 4.24\%) without any change to the model.} 
%
%
%
%
\paragraph{Marginals}
We now consider the effect of individual constraints by marginalizing out the \rateconstraintshort{} (Figure~\ref{fig:marginals_both}-Left) and \depthconstraintshort{} (Figure~\ref{fig:marginals_both}-Right). By examining the former, we aim to answer two specific questions: what is the effect of our \depthconstraintshort{} range on the ordinal results of the \metricnameshort{} metric, and what is the effect of the \depthconstraintshort{} on performance? For the former question, we see that our method is unambiguously better---that is, best or within one standard error of best at all \depthconstraintshort{}---on both tasks. However, the improved performance of our method on the VOT task is primarily due to its ability to effectively handle greater \depthconstraintshort{}s: if our evaluation were limited to $\text{\depthconstraintshort{}} \leq 2$, the \metricnameshort{} would be within one standard error of the Smart Replacement and Mean baselines.

Further consideration of the interaction between \depthconstraintshort{} and mean error provides meaningful insight into both our method and the findings of previous work. First, while other methods begin to meaningfully degrade as \depthconstraintshort{} increases, our method does not show such severe trends. Next, the \depthconstraintshort{} of one used in previous work~\cite{lemmer_crowdsourcing_2021,uehara_learning_2022} has meaningful shortcomings: all aggregation functions, with the exception of Mean, are improved by increasing the \depthconstraintshort{} beyond one on the referring expression comprehension task and, on the video object tracking task, the finding of previous work~\cite{lemmer_crowdsourcing_2021} that Smart Replacement is significantly better than Naive Replacement is only supported at the \depthconstraintshort{} of one used in their evaluation.
%
%

When the \depthconstraintshort{} is marginalized, our method is best or within one standard error of best at all \rateconstraintshort{}s. Broadly speaking, the behavior of this marginal is as expected: error decreases as \rateconstraintshort{} increases for all aggregation functions with the exception of naive replacement, which increases at higher \rateconstraintshort{}s. As noted when naive replacement was first used as a baseline~\cite{lemmer_crowdsourcing_2021}, this is due to the tendency to defer correct inferences at higher \rateconstraintshort{}s and replace them with potentially low-quality human inputs. 

\section{Discussion}
%
%
\paragraph{Alternate Task Definitions}
While our work is motivated by scenarios where human-provided information is used to define the task, there are two other hazy oracle formulations that should be considered: for applications such as Keypoint-Conditioned Viewpoint Estimation~\cite{szeto_click_2017} where the task could be performed without human information, the model could choose to solicit the hazy oracle only when a human-provided keypoint could change the answer---which is often not the case~\cite{lemmer_ground-truth_2021}. The task could also be defined by an automated hazy oracle: for example, Gurari \etal{} show that different segmentation algorithms work better in different cases~\cite{gurari_pull_2016}, which would have implications for a subsequent Video Object Segmentation~\cite{perazzi_benchmark_2016}. Our evaluation and solution extend trivially to both cases.
%
%
\paragraph{Deployment}
When shifting from dataset-based evaluation to practical human interaction, a few additional factors must be considered. First, we must define a threshold on the deferral score that targets an error or deferral rate. This requires an additional pass on a validation set and may use a method analogous to Selective Guaranteed Risk~\cite{geifman_selective_2017} if performance guarantees are required. Second, the datasets used for the experiments in this work were crowdsourced meaning that, although our findings are directly applicable to a crowdsourcing setting, factors that are dependent on an individual human are not evaluated: individuals have different biases and variances and the quality of the provided inputs may be dependent on the number of deferrals that have occurred. Works motivated by interaction with individuals should consider this during dataset procurement and evaluation.

\section{Conclusion}
Despite the intuitive benefit of deferred inference when information is provided by hazy oracles, previous works have performed only surface-level analyses under limited experiments, leading to an inability to effectively develop and compare methods. Through formalization of deferred inference, a novel evaluation metric, and demonstration of a straightforward method that provides meaningful improvement across two disparate applications, we hope to enable and motivate further research into this impactful problem.

\section{Acknowledgments} Toyota Research Institute (“TRI”) provided funds to assist the authors with their research but this article solely reflects the opinions and conclusions of its authors and not TRI or any other Toyota entity. This work was also partially supported by NSF CNS 1628987, the Google Faculty Research Award, and a gift from NEC.
\fontsize{9.1pt}{10.1pt} \selectfont
\bibliography{references}
\end{document}


\maketitle
\section*{DBSCAN Parameter Search}
Our implementation of deferred inference on the single-target VOT application relies on expectation maximization (EM)~\cite{dempster_maximum_1977} to produce distributions from the stochastic forward passes. To determine the number of clusters with which to initialize EM, we use DBSCAN~\cite{ester_density-based_1996}. DBSCAN relies on two hyperparameters: $\epsilon$ and \textit{minimum samples}. We performed a gridsearch across these two parameters---shown in Table~\ref{tab:dbscan-gridsearch}---to determine which to use for our experiments.
\begin{table}[h!]
\centering
\includegraphics{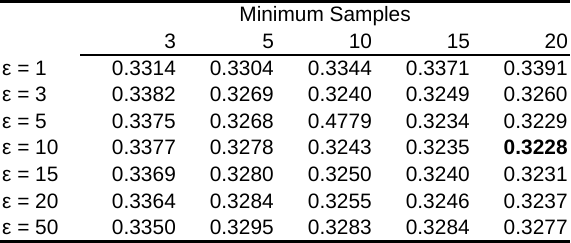}
\caption{The effect of DBSCAN parameters on the deferral-free mean error (1-IoU) of our method. Lower is better.}
\label{tab:dbscan-gridsearch}
\end{table}
\newpage
\section*{Additional Marginals}
We show the marginals for all three splits (val, testA, testB) of the referring expression comprehension task to augment the val split marginal shown in the main text. Across all three splits, the conclusions drawn from inspecting the validation split are largely true, though our method does show some small performance degradation at higher DDCs on the testB split. 
\begin{figure}[h!]
\centering
\includegraphics[width=\linewidth]{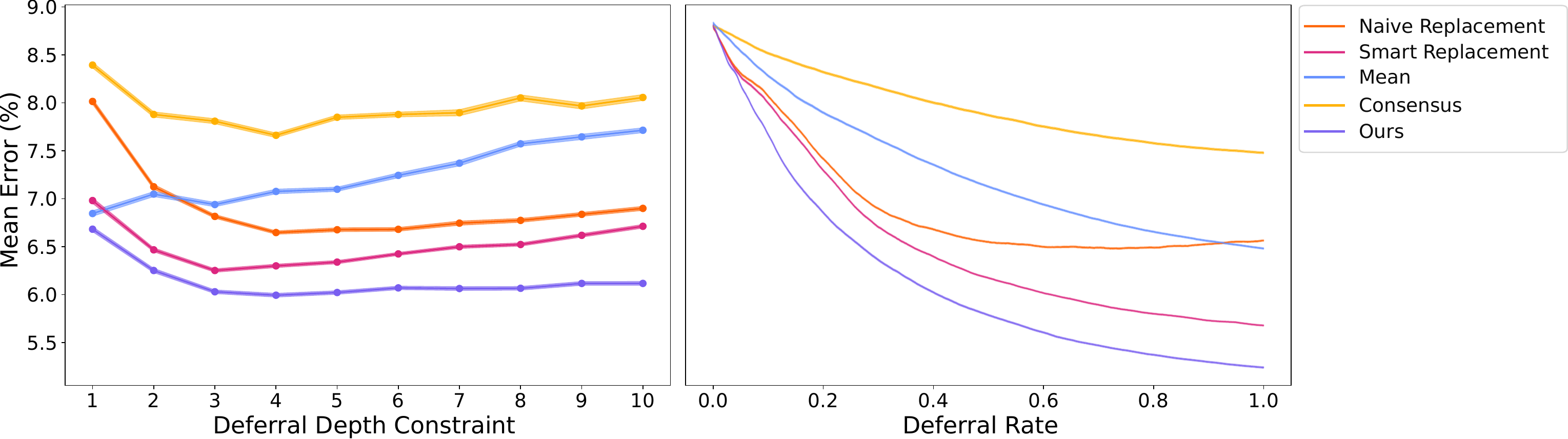}
\caption{Marginals for the Validation Split.}
\label{fig:marginal-val}
\end{figure}
\begin{figure}[h!]
\centering
\includegraphics[width=\linewidth]{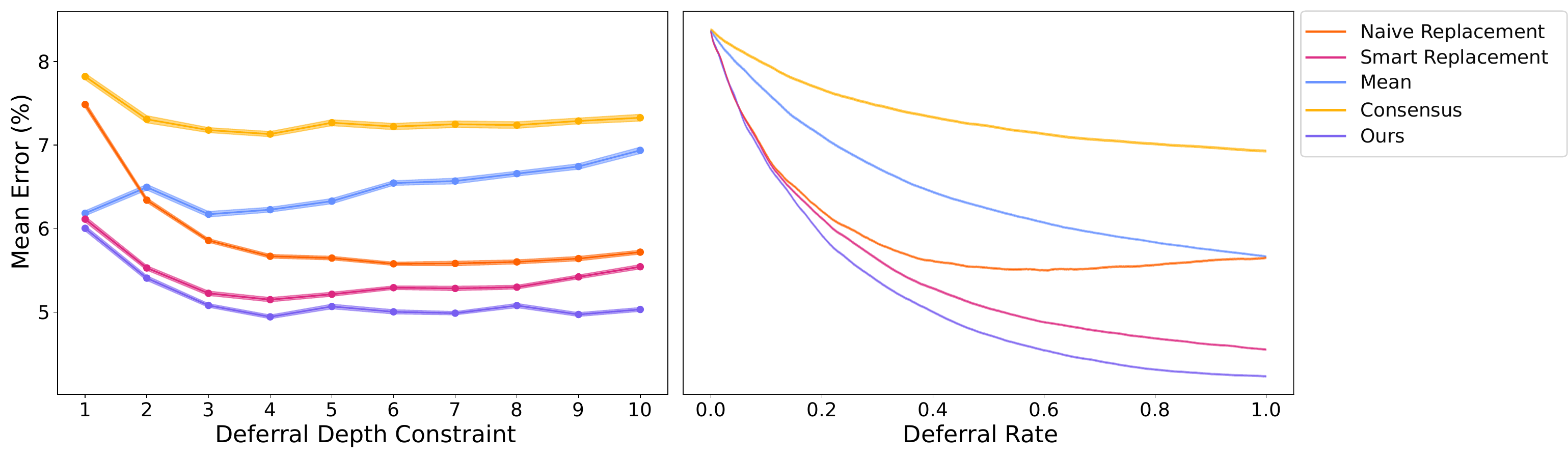}
\caption{Marginals for the TestA Split.}
\label{fig:marginal-testA}
\end{figure}
\begin{figure}[h!]
\centering
\includegraphics[width=\linewidth]{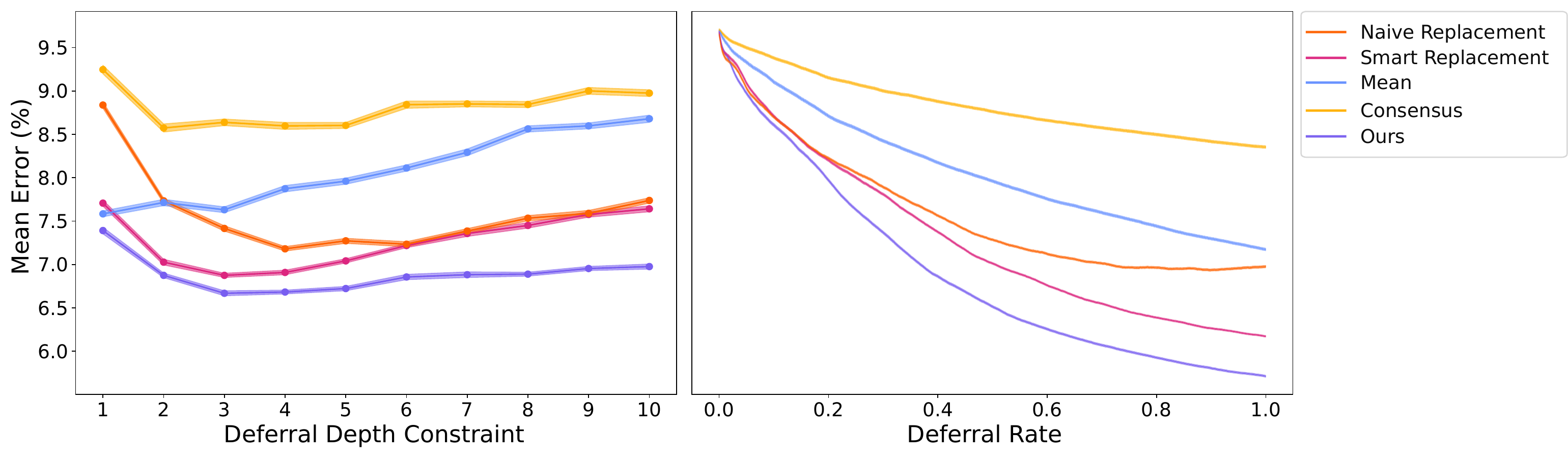}
\caption{Marginals for the TestB Split.}
\label{fig:marginal-testB}
\end{figure}
\newpage
\section*{The Importance of Dropout}
All experiments presented in the main text used Monte Carlo dropout~\cite{gal_dropout_2016} to produce (VOT) or improve the veracity of (RefExp) the task model's output distribution. In the latter case, we can compare performance when MC dropout is and isn't enabled. We show this in Table~\ref{tab:dropout_effect}: all conditions improve over deferral-free inference, but conditions with MC dropout enabled perform better. This improvement is particularly large for our method: since we do not discard information between human inputs, error in the underlying belief compounds over time.
\begin{table}[h!]
\centering
\includegraphics[width=\linewidth]{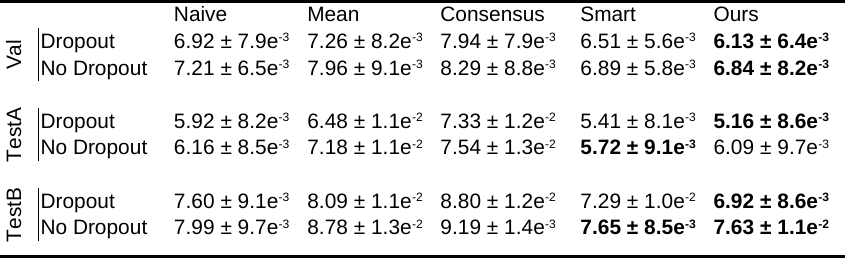}
\caption{The DEV with and without dropout for all selection functions. Best selection function for split/MC Dropout condition in bold.}
\label{tab:dropout_effect}
\end{table}
\section*{Per-Task Expression Counts}
To motivate our use of the RefCOCO dataset, we count the number of referring expressions for every individual task in the three most commonly used datasets: RefCOCO, RefCOCO+~\cite{kazemzadeh_referitgame_2014} and RefCOCOg~\cite{mao_generation_2016}. We see that a substantial number of tasks (objects) in the RefCOCO+ and RefCOCOg datasets have only one referring expression, making evaluations of deferral methods on these datasets less meaningful.
\begin{table}[h!]
\centering
\includegraphics[width=\linewidth]{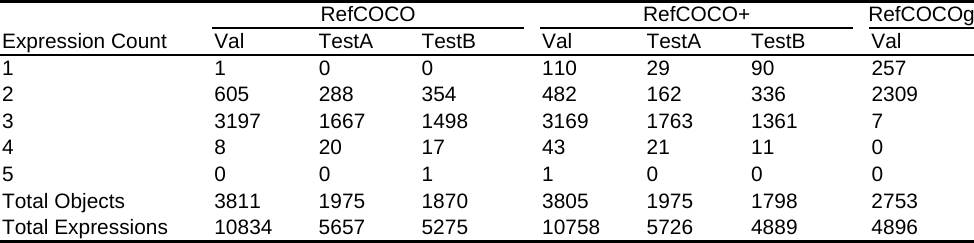}
\caption{Number of expressions per task (target object) for three common referring expression comprehension datasets.}
\label{fig:expression counts}
\end{table}

\newpage
\section*{Text Embedding Perturbation}
\begin{figure}[h!]
\centering
\includegraphics[width=\linewidth]{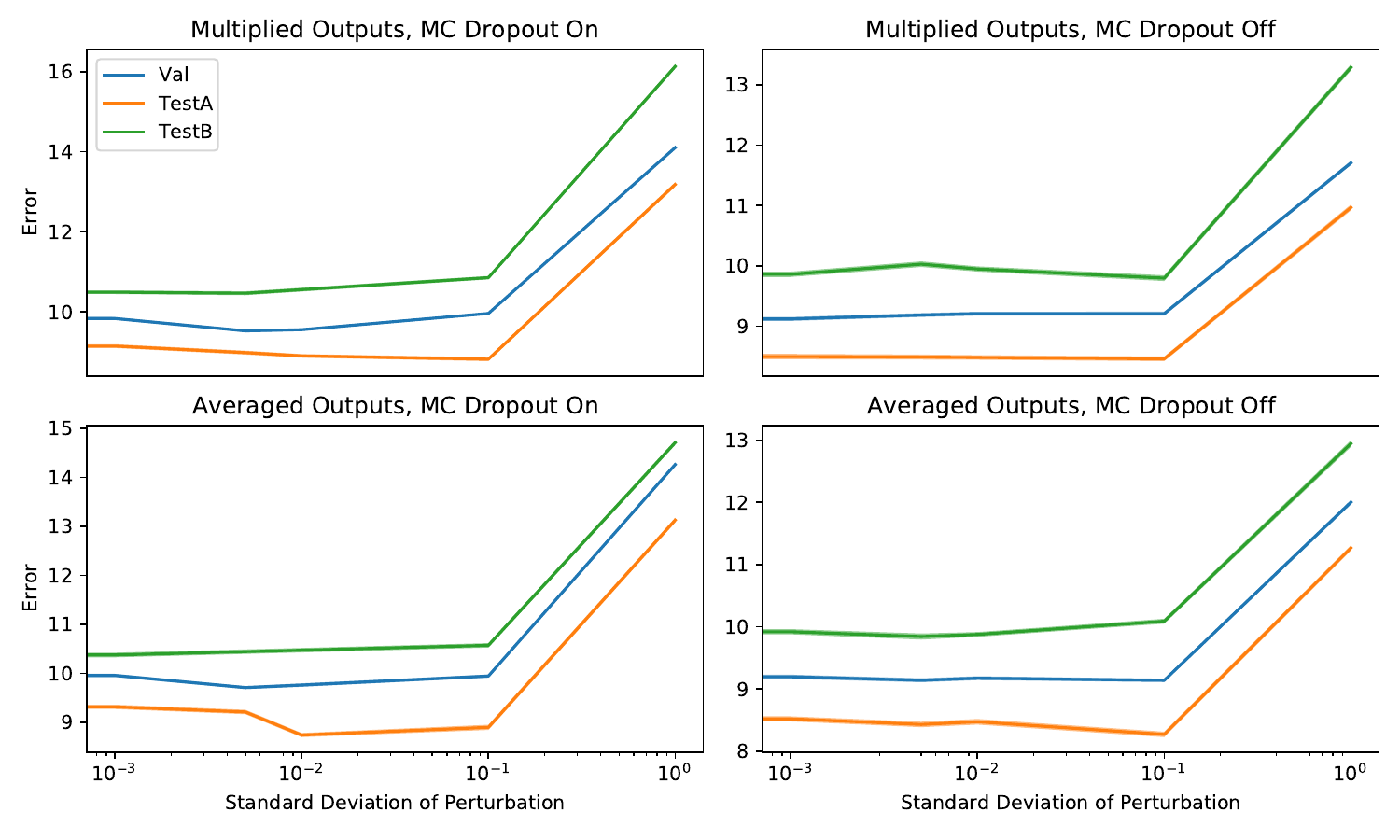}
\caption{Effect of perturbing text embeddings on error.}
\label{fig:wobblies}
\end{figure}
We further motivate the use of deferred inference by demonstrating that the accuracy improvements can not be matched by automated perturbations. To do this, we add random Gaussian noise of various standard deviations to the UNITER~\cite{chen_uniter_2020} text embeddings for the referring expression comprehension application. For every task we draw 5,000 samples with the same referring expression and combine the output distributions either as the average (averaged outputs) or the product (multiplied outputs). The error under all four conditions is shown in Figure~\ref{fig:wobblies}. As we see, any error reduction is much less than what is enabled by deferred inference, and the additional user burden is justified.

\newpage
\bibliography{references}